\documentclass{article}

\usepackage[preprint]{neurips_2025}

\usepackage{graphicx}
\usepackage{amsmath}

\usepackage[utf8]{inputenc} 
\usepackage{dirtytalk}
\usepackage[T1]{fontenc}    
\usepackage{hyperref}       
\usepackage{url}            
\usepackage{booktabs}       
\usepackage{amsfonts}       
\usepackage{nicefrac}       
\usepackage{microtype}      
\usepackage{xcolor}         
\usepackage[most]{tcolorbox}
\definecolor{Hpurple}{HTML}{69005f}
\usepackage{tikz}
\usepackage{lipsum}
\usepackage{helvet}
\tcbset{
  llmprompt/.style={
    colback=gray!5,
    colframe=black!50,
    boxrule=0.5pt,
    arc=0pt,
    outer arc=0pt,
    boxsep=4pt,
    left=4pt,
    right=4pt,
    top=4pt,
    bottom=4pt,
    fontupper=\ttfamily\small,
  }
}

\usepackage{listings}
\lstdefinestyle{mypython}{
  language=Python,
  basicstyle=\ttfamily\small,
  keywordstyle=\color{blue},
  stringstyle=\color{green!50!black},
  commentstyle=\color{gray}\itshape,
  numbers=none,   
  frame=none,     
  breaklines=true, breakatwhitespace=true,
  tabsize=4,
}

\title{Automated scientific minimization of regret}

\author{%
  Marcel Binz \\
  Helmholtz Munich\\
  Institute for Human-Centered AI \\
  \texttt{marcel.binz@helmholtz-munich.de} \\
  \And
  Akshay K. Jagadish \\
  Helmholtz Munich\\
  Institute for Human-Centered AI \\
  \texttt{akshay.jagadish@helmholtz-munich.de} \\
  \And
  Milena Rmus \\
  Helmholtz Munich\\
  Institute for Human-Centered AI \\
  \texttt{milena.rmus@helmholtz-munich.de} \\
  \And
  Eric Schulz \\
  Helmholtz Munich\\
  Institute for Human-Centered AI \\
  \texttt{eric.schulz@helmholtz-munich.de} \\
}

\begin{document}

\maketitle

\begin{abstract}
    We introduce \emph{automated scientific minimization of regret} (ASMR) -- a framework for automated computational cognitive science. Building on the principles of scientific regret minimization, ASMR leverages Centaur -- a recently proposed foundation model of human cognition -- to identify gaps in an interpretable cognitive model. These gaps are then addressed through automated revisions generated by a language-based reasoning model. We demonstrate the utility of this approach in a multi-attribute decision-making task, showing that ASMR discovers cognitive models that predict human behavior at noise ceiling while retaining interpretability. Taken together, our results highlight the potential of ASMR to automate core components of the cognitive modeling pipeline.
\end{abstract}

\section*{Introduction}

The combination of large-scale behavioral data sets and advances in machine learning has enabled the development of highly predictive models of human behavior \citep{binz2024centaur, peterson2021using, eckstein2024hybrid}. Yet, while these models excel at predicting behavior, they offer limited insight into the underlying cognitive mechanisms. A central challenge, therefore, is how to move beyond prediction and leverage these models to improve our understanding of human cognition.

A promising framework for achieving this goal is \emph{scientific regret minimization} \citep{agrawal2020scaling}. It takes a black-box predictive model and compares it against an interpretable cognitive model on a per-data-point basis. This comparison is used to identify data points that are, in principle, predictable -- because they are correctly predicted by the black-box model --  but are not yet captured by the cognitive model. Patterns in these data points are analyzed and incorporated back into the cognitive model through an iterative refinement process.

Two factors have prevented the broader adoption of this framework: (1) it requires large data sets to train a predictive model, and (2) identifying patterns in the resulting data points can be challenging. The first challenge is typically addressed by conducting a large-scale study within the experimental paradigm of interest and training a black-box model -- typically some form of neural network -- on the resulting data. This approach imposes substantial overhead, thereby limiting the scope of scientific regret minimization. The second challenge remains largely unresolved and is typically addressed through what might be called the \say{method of staring} where researchers manually inspect the identified data points until a recognizable pattern is recognized that can be incorporated into the cognitive model.

The present paper introduces the idea of \emph{automated scientific minimization of regret} (ASMR) -- a framework that offers a solution to both of these issues. ASMR relies on Centaur -- a foundation model of human cognition -- as the predictive model \citep{binz2024centaur}. Because Centaur was trained on a large collection of behavioral experiments, it can predict human behavior across domains without requiring additional data collection or task-specific training. To address the second issue, ASMR uses state-of-the-art reasoning models such as DeepSeek-R1 \citep{deepseekai2025deepseekr1incentivizingreasoningcapability} or Qwen3 \citep{qwen3}. Leveraging their language-based reasoning capabilities, these models can analyze failure modes and suggest new candidate models in an automated fashion and without the need for human intervention.

\begin{figure}
    \centering
    \includegraphics[width=1.03\textwidth]{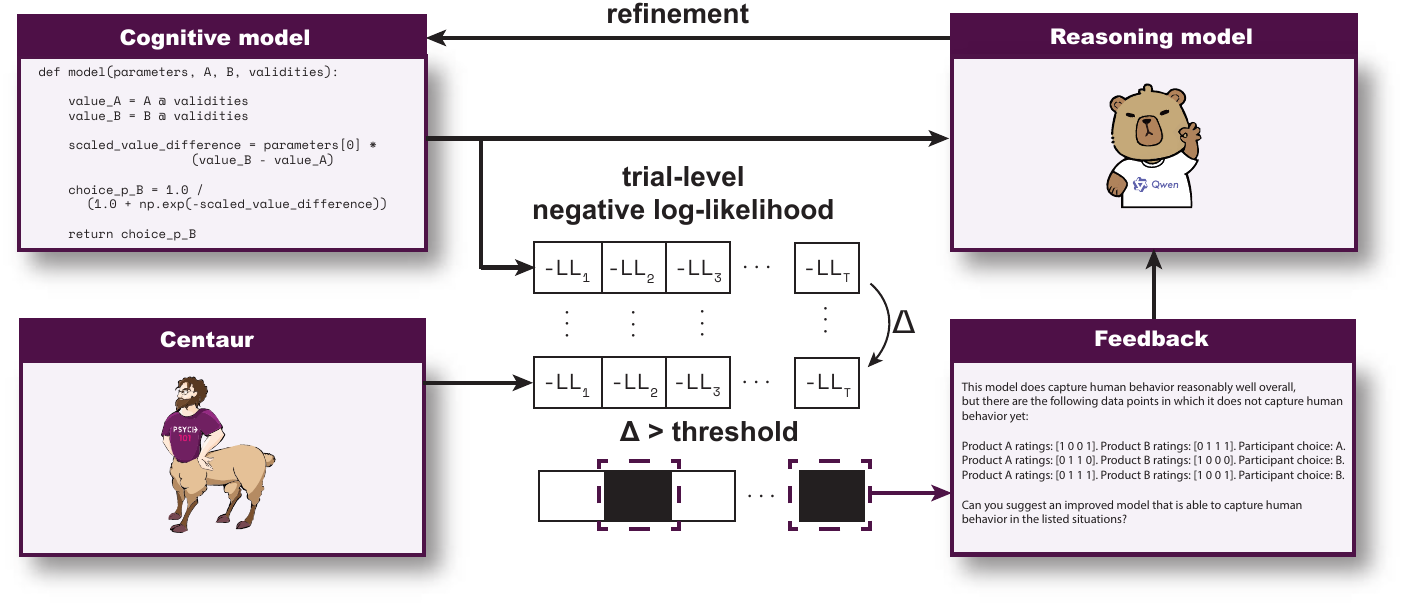}
    \caption{ASMR pipeline. Centaur is used as a guide to identify gaps in an interpretable cognitive model, resulting in a set of data points that are, in principle, predictable but are not currently accounted for. These points, together with the model's code and a brief instruction, are then provided to a language-based reasoning model. The reasoning model generates modifications to the cognitive model -- a process which can be iterated multiple times.}
    \label{fig:1}
\end{figure}

We present a case study illustrating the potential of ASMR. Without any human input, ASMR discovers cognitive models that match Centaur in predictive performance while retaining interpretability. Taken together, these results demonstrate the feasibility of fully automated scientific discovery  \citep{musslick2024closed, binz2025should, musslick2025automating, castro2025discovering, rmus2025towards} guided by large-scale predictive models.

\section*{Results}

We apply ASMR to a multi-attribute decision-making paradigm \citep{hilbig2014generalized}, in which participants made repeated judgments between two fictitious products, each rated by four experts. Each expert provides a binary rating for both products, either approving or disapproving of them. Participants completed 96 trials in which they indicated which product they believed to be of higher quality. To guide their decision-making process, they were furthermore provided with the validity of each expert, defined as the probability that the product an expert approves is indeed the objectively better one on those occasions where the expert's two ratings differ. We focus our analysis on a subset of participants that are not part of Centaur's training data. 

ASMR begins by extracting log-likelihoods of human responses for each participant and trial based on Centaur's predictions. In each iteration, it compares the resulting log-likelihoods to those of the cognitive model under evaluation. Free parameters of the cognitive model are fitted to the data using standard maximum likelihood estimation. Data points where the difference in log-likelihood exceeds a predefined threshold are identified and, along with the cognitive model code and a set of natural language instructions, passed to Qwen3-32B -- a state-of-the-art reasoning model \citep{qwen3}. Qwen3-32B then modifies the cognitive model with the goal of improving its predictions on the identified data points (see Figure 1). This process repeats for multiple iterations. In our experiments, five iterations were sufficient, though the optimal number may vary depending on task complexity.

We initialize the cognitive model using three strategies frequently employed in decision-making literature: the take-the-best heuristic, an equal weighting heuristic, and a weighted additive strategy \citep{gigerenzer2000simple, binz2022heuristics}. Reported results are averaged over ten simulations per model class and final model performance is evaluated using the Akaike information criterion (AIC). 

\begin{figure}
    \textbf{a} \hspace{6.7cm} \textbf{b} \\
    \vspace{-0.3cm}
    \includegraphics[]{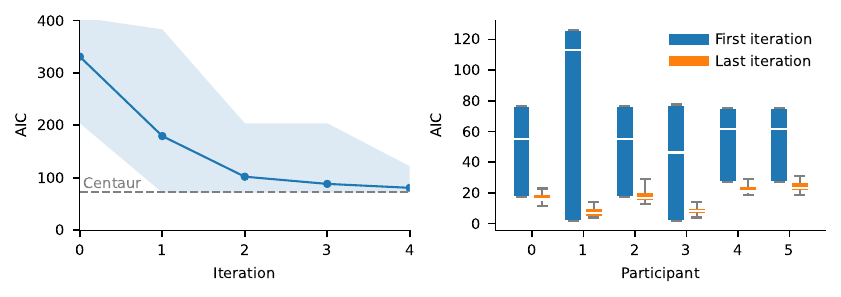} 
    
    \textbf{c}
    
    \begin{lstlisting}[style=mypython]
def model(parameters, option_A, option_B):
    # define feature validity weights (importance of each feature)
    validities = np.array([parameters[0] * 0.9, 0.8, 0.7, 0.6])
    
    # compute weighted value for each option
    value_A = option_A @ validities
    value_B = option_B @ validities

    # compute scaled difference in value
    scale_value_difference = parameters[1] * (value_B - value_A)

    # apply logistic function to obtain choice probabilities
    choice_probability_B = 1.0 / (1.0 + np.exp(-scale_value_difference))

    return choice_probability_B
\end{lstlisting}
    
    \caption{Summary of results. \textbf{a}, Improvement of aggregated AIC scores across iterations of ASMR. The solid line show the average AIC score, while the shaded areas represent the worst and best model at a given iteration. \textbf{b}, AIC scores from the models at the first and last iteration for each individual participant. \textbf{c}, Python code for one of the discovered models with the lowest AIC score.}
    \label{fig:2}
\end{figure}

We found that ASMR  rapidly improves the initial model. Within five iterations, the discovered models reach an average AIC score (M = 80.72, SD = 17.05) that approaches Centaur’s goodness-of-fit (AIC$_\text{Centaur} = 72.5$; see Figure \ref{fig:2}a). Notably, the best-performing model (AIC$_\text{ASMR} = 71.73$) even surpasses Centaur and matches the performance of prior approaches \citep{rmus2025towards}. Furthermore, ASMR consistently improves the AIC score of every participant, as shown in Figure \ref{fig:2}b. Figure~\ref{fig:2}c presents one of the discovered models with the lowest AIC score. This example illustrates an adaptive upweighting of the highest-validity expert, enabling interpolation between take-the-best and weighted-additive strategies \citep{parpart2018heuristics} -- a pattern that consistently emerges across multiple discovered models.

\section*{Discussion}

We have presented ASMR as a framework for the automated discovery of interpretable cognitive models. ASMR invokes Centaur -- a foundation model of human cognition -- as a reference model to reveal data points that are, in principle, predictable but are not captured by a given cognitive model. A state-of-the-art reasoning model is then prompted with this information and tasked with revising the cognitive model. When applied iteratively, this process enables the discovery of novel, interpretable cognitive theories.

While the results presented in this article serve as a first proof-of-concept, they open up several avenues for future exploration. A central question for future work concerns the nature of the feedback signal. We have shown that simply presenting negative data points is sufficient for discovering novel cognitive models. However, in more complex domains, it may be necessary to supplement this with additional forms of feedback, such as positive examples, or performance metrics. In our simulations, the reasoning model was furthermore only given the most recent version of the cognitive model for revision. Improvements in context window size may eventually lift this constraint and allow reasoning models to process and revise multiple proposals within a single prompt. Fully evaluating these alternatives necessitates a benchmark for automated scientific discovery in the cognitive sciences. Ideally, such a benchmark should span multiple experimental paradigms and cover a representative spectrum of cognitive modeling scenarios.

Ultimately, we expect that ASMR will enable the rapid generation of a wide range of cognitive models. To manage this growing set, we envision a searchable database that systematically links cognitive models to the experimental data they aim to explain. This database could be queried to identify models relevant to specific tasks or populations and would provide a foundation for large-scale meta-analyses across studies, paradigms, and modeling approaches.

\newpage 

\section*{Methods}

We initialize ASMR with the following cognitive models:

\begin{lstlisting}[style=mypython]
# weighted addtive strategy
NUM_PARAMETERS = 1

def model(parameters, option_A, option_B):
    """
    Compute the probability of choosing Option B over Option A.

    Parameters
    ----------
    parameters : np.ndarray of shape (num_parameters,)
        Model parameters.

    option_A : np.ndarray of shape (num_trials, num_features)
        Feature matrix for Option A across trials.

    option_B : np.ndarray of shape (num_trials, num_features)
        Feature matrix for Option B across trials.

    Returns
    -------
    choice_probability_B : np.ndarray of shape (num_trials,)
        The predicted probability of choosing Option B on each trial.
    """

    # define feature validity weights (importance of each feature)
    validities = np.array([0.9, 0.8, 0.7, 0.6])
    
    # compute weighted value for each option
    value_A = option_A @ validities
    value_B = option_B @ validities

    # compute scaled difference in value
    scale_value_difference = parameters[0] * (value_B - value_A)

    # apply logistic function to obtain choice probabilities
    choice_probability_B = 1.0 / (1.0 + np.exp(-scale_value_difference))

    # clip probabilities to avoid numerical issues
    choice_probability_B = np.clip(choice_probability_B, 0.00001, 1 - 0.00001)

    return choice_probability_B
\end{lstlisting}

\begin{lstlisting}[style=mypython]
# take-the-best heuristic
NUM_PARAMETERS = 1

def model(parameters, option_A, option_B):
    """
    Compute the probability of choosing Option B over Option A.

    Parameters
    ----------
    parameters : np.ndarray of shape (num_parameters,)
        Model parameters.

    option_A : np.ndarray of shape (num_trials, num_features)
        Feature matrix for Option A across trials.

    option_B : np.ndarray of shape (num_trials, num_features)
        Feature matrix for Option B across trials.

    Returns
    -------
    choice_probability_B : np.ndarray of shape (num_trials,)
        The predicted probability of choosing Option B on each trial.
    """

    # define feature validity weights (importance of each feature)
    validities = np.array([1.0, 0.5, 0.25, 0.125])
    
    # compute weighted value for each option
    value_A = option_A @ validities
    value_B = option_B @ validities

    # compute scaled difference in value
    scale_value_difference = parameters[0] * (value_B - value_A)

    # apply logistic function to obtain choice probabilities
    choice_probability_B = 1.0 / (1.0 + np.exp(-scale_value_difference))

    # clip probabilities to avoid numerical issues
    choice_probability_B = np.clip(choice_probability_B, 0.00001, 1 - 0.00001)

    return choice_probability_B
\end{lstlisting}

\begin{lstlisting}[style=mypython]
# equal weighting heuristic
NUM_PARAMETERS = 1

def model(parameters, option_A, option_B):
    """
    Compute the probability of choosing Option B over Option A.

    Parameters
    ----------
    parameters : np.ndarray of shape (num_parameters,)
        Model parameters.

    option_A : np.ndarray of shape (num_trials, num_features)
        Feature matrix for Option A across trials.

    option_B : np.ndarray of shape (num_trials, num_features)
        Feature matrix for Option B across trials.

    Returns
    -------
    choice_probability_B : np.ndarray of shape (num_trials,)
        The predicted probability of choosing Option B on each trial.
    """

    # compute weighted value for each option
    value_A = option_A.sum(-1)
    value_B = option_B.sum(-1)

    # compute scaled difference in value
    scale_value_difference = parameters[0] * (value_B - value_A)

    # apply logistic function to obtain choice probabilities
    choice_probability_B = 1.0 / (1.0 + np.exp(-scale_value_difference))

    # clip probabilities to avoid numerical issues
    choice_probability_B = np.clip(choice_probability_B, 0.00001, 1 - 0.00001)

    return choice_probability_B
\end{lstlisting}

In each iteration, the cognitive model is fitted to human responses on a per-subject basis using maximum likelihood estimation. This procedure is implemented using \textsc{scipy}'s minimize function with the BFGS algorithm. We then subtract the cached negative log-likelihoods obtained from Centaur from those of the fitted cognitive model. Next, we submit all data points where this difference exceeds a threshold of $\Delta \geq 0.05$ -- together with the model's code and a brief instruction -- to Qwen3-32B \citep{qwen3}. For this, we rely on the four-bit quantized version of the \textsc{unsloth} package with sampling parameters set to the recommended values \citep{qwen3}. We use the following prompt template:

\begin{tcolorbox}[llmprompt]
I am studying human behavior in a multi-attribute decision-making experiment.

In this experiment, participants encounter a number of trials, in which they have to choose between two options labelled A and B.

These options are fictitious products that are each characterized by four features.

Each feature corresponds to a binary rating of an expert, either approving of the product (1) or not (0).

The four experts are ordered based on their validity (taking values of 90\%, 80\%, 70\%, and 60\%), with the first feature corresponding to the ratings from the highest validity expert.

In each trial, people have to predict which of the shown options is superior in terms of quality based on the presented information.

I have the following computational model that is currently my best guess for how people make decisions in this experiment: \\

[INSERT MODEL CODE] \\

This model does capture human behavior reasonably well overall, but there are the following data points in which it does not capture human behavior yet: \\

[INSERT DATA POINTS] \\

Can you suggest an improved model that is able to capture human behavior in the listed situations? \\

Please structure your answer as follows:

* Keep the structure of the function exactly the same.
* Do not change the docstring.

* State the number of free parameters before the model function using the NUM\_PARAMETERS variable.

* Do not write any text besides that and do not elaborate any further.
\end{tcolorbox}

Finally, the cognitive model is updated based on Qwen3's output, and the entire process is repeated. We ran ten simulations per model class, each with five iterations, and report results averaged across  simulations unless otherwise noted. ASMR -- in its current form -- requires access to a single 80GB GPU (e.g., A100) but could be modified to run on fewer computational resources by using smaller models.

\newpage

\bibliographystyle{unsrtnat}

\bibliography{main}

\end{document}